\documentclass[letterpaper, 10 pt, conference]{ieeeconf}  

\IEEEoverridecommandlockouts                              

\usepackage{amsmath}
\usepackage{amssymb}
\usepackage{hyperref}
\usepackage{graphicx}
\usepackage{makecell}
\usepackage{leftindex}
\usepackage{subcaption}
\usepackage{svg}
\usepackage{booktabs} 
\usepackage{bm}

\newcommand\state{\bm{x}}
\newcommand\control{\bm{u}}

\newcommand\R{\mathbb{R}}
\newcommand\q{\bm{q}}
\newcommand\qdot{\bm{\dot q}}
\newcommand\qddot{\bm{\ddot q}}
\newcommand\force{\bm{\lambda}}

\title{\LARGE \bf
Learning-Guided Force-Feedback Model Predictive Control \\ with Obstacle Avoidance for Robotic Deburring
}

\author{Krzysztof Wojciechowski$^{1}$, Ege Gursoy$^{1}$, Arthur Haffemayer$^{1}$, Sebastien Kleff$^{2}$, \\
Vincent Bonnet$^{1,3}$, Florent Lamiraux$^{1}$ and Nicolas Mansard$^{1,4}$ 
\thanks{This work is supported by the European project AGIMUS (under GA no.101070165), ANITI (ANR-19-P3IA-0004), NERL (ANR-23-CE94-0004-02), AS2 (ANR-22-EXOD-0006) PEPR O2R project and by Défi Clé Robotique centrée sur l'Humain funded by Région Occitanie, France.}
\thanks{$^{1}$LAAS-CNRS, Universite de Toulouse, CNRS, Toulouse, France}%
\thanks{$^{2}$Inria, AUCTUS team, Talence, France}%
\thanks{$^{3}$Image and Pervasive Access Laboratory (IPAL), CNRS-UMI, 2955, Singapore}%
\thanks{$^{4}$Artificial and Natural Intelligence Toulouse Institute, France}%
}

\begin{document}

\maketitle
\thispagestyle{empty}
\pagestyle{empty}

\begin{abstract}
Model Predictive Control (MPC) is widely used for torque-controlled robots, but classical formulations often neglect real-time force feedback and struggle with contact-rich industrial tasks under collision constraints. Deburring in particular requires precise tool insertion, stable force regulation, and collision-free circular motions in challenging configurations, which exceeds the capability of standard MPC pipelines. We propose a framework that integrates force-feedback MPC with diffusion-based motion priors to address these challenges. The diffusion model serves as a memory of motion strategies, providing robust initialization and adaptation across multiple task instances, while MPC ensures safe execution with explicit force tracking, torque feasibility, and collision avoidance. We validate our approach on a torque-controlled manipulator performing industrial deburring tasks. Experiments demonstrate reliable tool insertion, accurate normal force tracking, and circular deburring motions even in hard-to-reach configurations and under obstacle constraints. To our knowledge, this is the first integration of diffusion motion priors with force-feedback MPC for collision-aware, contact-rich industrial tasks.
\end{abstract}

\section{Introduction}

Model Predictive Control (MPC) has long been the staple for motion generation on torque-controlled robots, with applications ranging from locomotion \cite{grandia_feedback_2019, romualdi_online_2022, dantec_whole-body_2022} to collision avoidance \cite{gaertner_collision-free_2021,haffemayer_collision_2025}. 
However, most MPC formulations optimize motion alone without explicitly incorporating real-time force feedback~\cite{Farshidian2017, Grandia2023, Mastalli2023}, making force tracking purely feedforward and sensitive to modeling errors.

In contrast, the control of contact forces has been a longstanding topic in robotics~\cite{Whitney1977ForceFC, Villani2008}, originally developed through direct and indirect feedback methods such as hybrid force/motion control~\cite{Mason1981}, impedance control~\cite{Hogan1985}, or admittance control~\cite{Whitney1977ForceFC}. These techniques offer reactivity and robustness but lack planning capabilities, which limits effectiveness in complex and contact-rich tasks. 

To bridge this gap some works attempted to incorporate optimization into force control, for instance through MPC formulations based on linear spring models~\cite{Killpack2016, Matschek2017, Husmann2019, Muller2019}, bounded admittance behavior~\cite{Wahrburg2016}, or adaptive impedance models~\cite{Pankert2020, Kazim2018, Minniti2021}. These approaches improve motion–force coordination but often rely on simplified or low-dimensional dynamics.

Conversely, another line of work attempted to extend MPC with force feedback, either by relaxing the rigid contact model~\cite{Neunert2018, Fahmi2020}, using spring models~\cite{gold_model_2023}, or modifying the control structure~\cite{Kleff22, Gazar21}. These hybrid approaches do demonstrate the importance of contact force prediction, but often face trade-offs: simplified dynamics~\cite{gold_model_2023}, low planning rates~\cite{wijayarathne_real-time_2023}, or increased state size and model approximations~\cite{Selvaggio2023}. Force feedback is typically either ignored or estimated from joint torques. More recently,~\cite{kleff:hal-04572399} proposed a unified formulation that combines force control and nonlinear MPC without these compromises. 


Despite progress, running such controllers online in real time remains challenging, especially for tasks like deburring, where stable contact and precise tool motions must be maintained under uncertainty \cite{makulavicius_industrial_2023}. In these tasks, not only the force magnitude but also the spatiotemporal profile of force application is critical. The robot must retain and reproduce task-specific motion strategies, for instance characteristic insertion poses and trajectories for each hole to deburr. While MPC typically exploits the previous solution as a warm start, this only accelerates convergence locally and does not provide a true memory of task-specific motion strategies across different instances.

\begin{figure}[!t]
    \centering
   \includegraphics[width=0.8\linewidth]{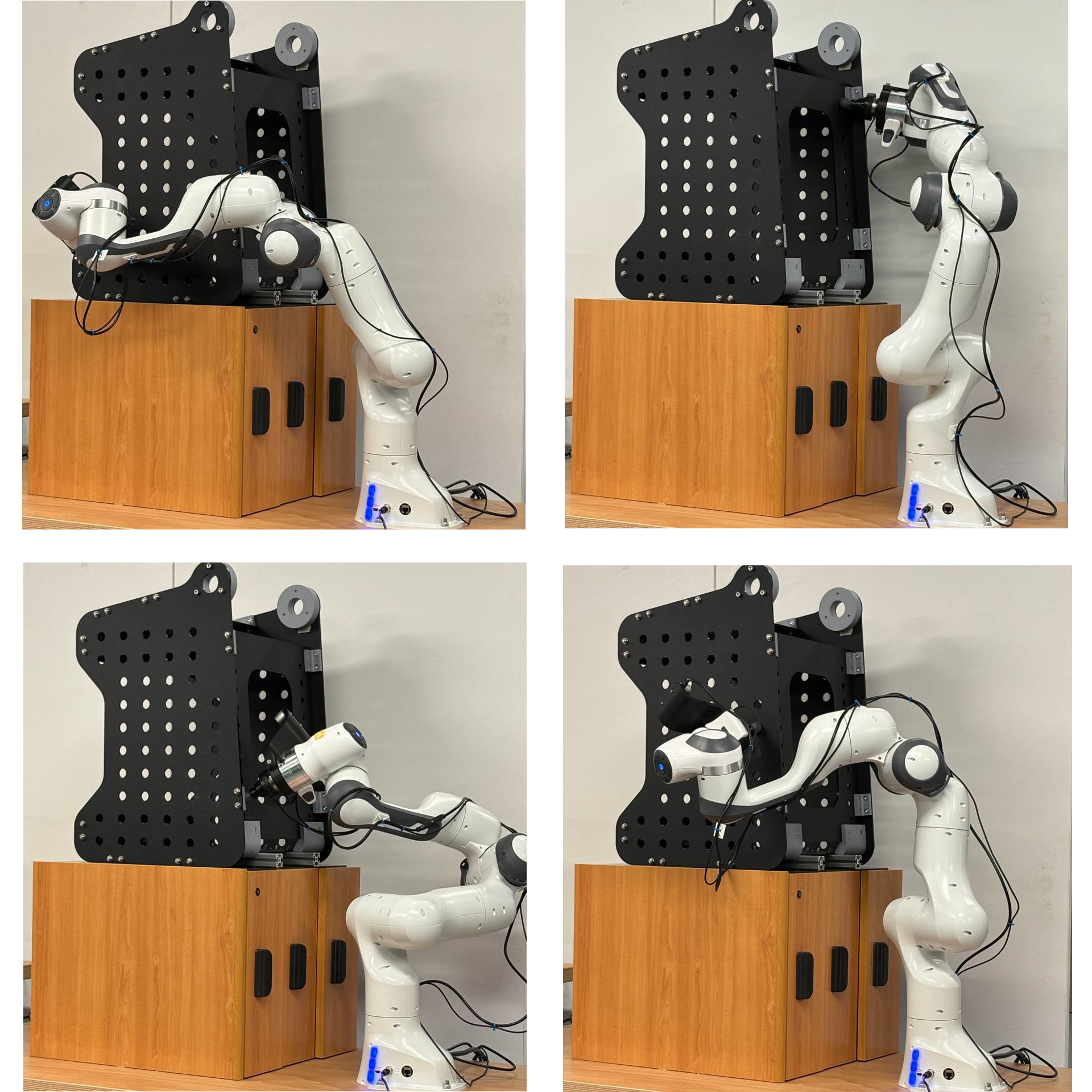}
   \setlength{\belowcaptionskip}{-20pt}
       \vspace{0pt}
        \caption{Torque-controlled Franka Emika Panda robot executing the deburring task in scenario 3 (see sec. \ref{sec:scenario2}). The robot inserts a custom end-effector into different holes of the workpiece and applies a constant normal force while performing circular motions, all while adapting to varying joint configurations and avoiding collisions with the environment.}  
        \label{fig:panda}
\end{figure}

To address this limitation, generative models have recently emerged as powerful motion priors. Classical representations such as dynamic movement primitives or Gaussian mixtures could encode demonstrations \cite{jetchev_trajectory_2009, mansard_using_2018, lembono_learning_2020, dantec_whole_2021}, but were limited in capturing the variability and multi-modality of real robot behavior. Diffusion models overcome these issues by learning rich distributions of feasible motions. Recent diffusion-based policies have shown that such models can generate diverse, feasible trajectories conditioned on goals or sensory input \cite{janner_planning_2022, zhou_diffusion_2024, chi_diffusion_2024}, making them a natural candidate for serving as a memory of motion in contact-rich planning.

\begin{figure}[!t]
    \centering
   \includegraphics[width=0.8\linewidth]{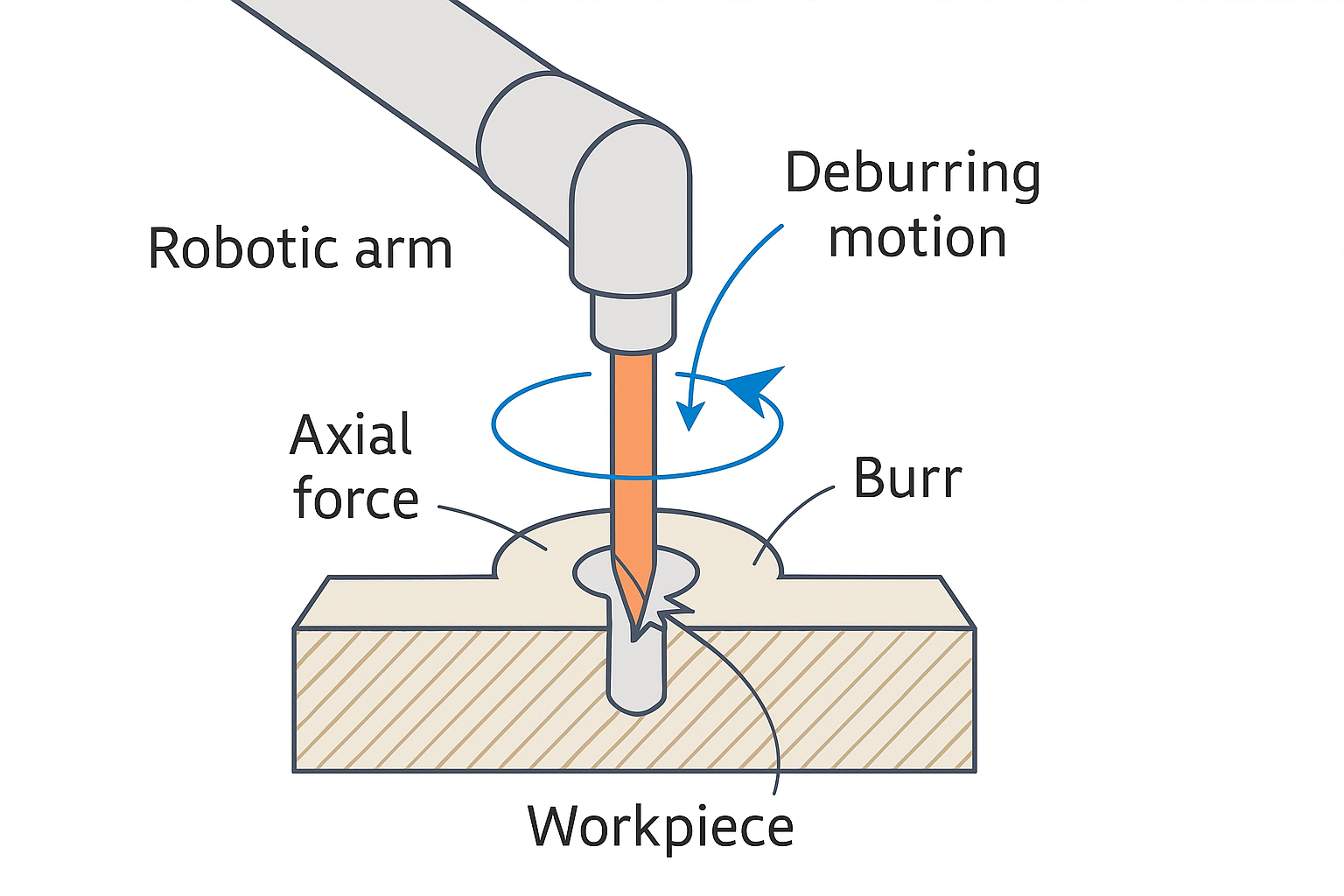}
   \setlength{\belowcaptionskip}{-20pt}
       \vspace{0pt}
        \caption{Robotic deburring process: the manipulator applies a constant normal force along the tool axis while performing a circular motion inside the hole to remove burrs with limited tangential forces.}  
        \label{fig:deburring}
\end{figure}
In this paper, we propose a framework that combines force-feedback MPC with diffusion-based motion priors. Diffusion models provide robust initialization and adaptation for the predictive controller, while MPC ensures safe execution with force regulation and collision avoidance. We demonstrate the approach on a deburring task that requires precise tool insertion, circular force-controlled motions, and obstacle avoidance. 

Our contributions are threefold. First, we formalize the deburring task and formulate the corresponding Optimal Control Problem (OCP). Second, we integrate diffusion-based motion priors into a force-feedback MPC. Finally, we experimentally validate the framework on a torque-controlled manipulator. Our experimental setup is shown in Fig. \ref{fig:panda}. To the best of our knowledge, this is the first integration of diffusion-based motion priors with force-feedback MPC for a contact-rich task such as deburring.

\section{Problem Statement}

Deburring is the process of removing excess material to smooth the edges of a machined metal part, as shown in Fig.~\ref{fig:deburring}. To this end, we define it as the insertion of a precision tool into a hole, followed by the application of a constant normal force $\force_{\perp}$ along the hole axis while performing a circular motion inside it to remove excess material. Tangential forces, $\force_{\parallel}$, should not be zero but only limited. In hole deburring, the tangential force component provides the cutting and abrading action. Excessive tangential force, however, can cause slip-out. The objective is to track the desired normal force and reference trajectory, while bounding tangential forces within the friction cone within a safe envelope, i.e. to respect the friction cone $\|\force_{\perp}\|_2 \leq \alpha \force_{\parallel}$.  In our empirical observations on dry steel deburring, a friction coefficient of $\alpha = 0.6$ is adopted. This value is conservative, ensuring stable contact and preventing slip-out.
 
Reaching and deburring small holes under collision avoidance is particularly challenging in real manufacturing settings. Fig. \ref{fig:panda} shows the realistic aeronautics-inspired workpiece used in our study. On the shop floor, it is common to reposition or re-fixture either the workpiece or the robot base to improve manipulability for each hole or batch of holes, which is an approach that increases setup time, cost, and system complexity. A more scalable alternative is to plan from a single, fixed robot pose. However, this makes the motion-planning problem especially hard: such workpieces are highly nonconvex, with thin corridor-like features and tight clearances around fixtures, creating narrow passages where small path deviations can cause collision or loss of reachability.

Overall to perform a deburring task the following elements are required (i) tight geometric clearances around the workpiece and fixture, (ii) strict orientation constraints to align the tool axis with the hole axis, and (iii) force control that prevent slippage.  These requirements must be satisfied under robot dynamics and rate limits at high control frequencies  leaving little time for collision checking, Jacobian/Hessian updates, or replanning. As a result, the task stresses both planning and control: it couples geometric feasibility with contact stability and real-time computation, making it an exciting benchmark for MPC-based contact-aware motion planning. 


\section{Force-feedback Model Predictive Control}
We recall here the force-feedback MPC approach described in~\cite{kleff:hal-04572399}. 
\subsection{Optimal Control Problem}
The OCP is defined as:
\begin{equation}
\label{sqp_equation}
\begin{aligned}
 \underset{\underline{\state} , \underline{\control}}{\min} 
 & \quad \sum_{t = 0}^{T-1} \ell_t (\state_t, \control_t) + \ell_T (\state_T) \\
\textit{s.t.} \quad 
 & \state_{t+1} = f_t(\state_t,\control_t), \quad 0 \leq t < T, \\
 & c_t(\state_t,\control_t) \geq 0, \quad 0 \leq t < T, \\
 & c_T(\state_T) \geq 0.
\end{aligned}
\end{equation}
where $\underline{\state} = \{\state_0,...,\state_T\},~\underline{\control} = \{\control_0,...,\control_{T-1}\}$ are the state and control input trajectories, $T\in\mathbb{N}^*$ is the horizon length, $\ell_t, f_t, c_t$ are the running cost, dynamics and constraint at stage $t$ and $\ell_T, c_T$ are the terminal cost and constraint. 


\subsection{Classical MPC}
For torque-controlled manipulators, a common state-control choice is the joint position–velocity pair $(\q,\qdot)$ and the joint torque input $\bm{\tau}$~\cite{Farshidian2017, Grandia2023, Mastalli2023}. 
In this case, for a fully actuated $n$-DoF manipulator in contact with the environment, the forward dynamics are expressed as:
\begin{equation}
\label{eq:joint_acc}
\begin{aligned}
    \qddot(\q, \qdot, \force, \bm{\tau}) 
    &= M^{-1}(\q)\left(\bm{\tau} - h(\q,\qdot) + J(\q)^\top \force\right)
\end{aligned}
\end{equation}
where $M(\q)\in\R^{n\times n}$ is the generalized inertia matrix, $h(\q,\qdot)\in\R^n$ denotes gravity and Coriolis terms, $J(\q)\in\R^{m\times n}$ is the contact Jacobian,  $\force\in\R^m$ are the contact forces with $1 \leq m \leq 3$, and $\qddot$ is the vector of joint accelerations.
 

This formulation, under the rigid contact assumption~\cite{Budhiraja2018}, defines the \textit{classical} MPC setting. Contact forces can be introduced as \textit{feedforward} terms in the dynamics, but they are not directly regulated through explicit force measurement feedback. Therefore, force objectives can be encoded in the cost function, yet tracking remains model-dependent and sensitive to uncertainties.

\subsection{Force-feedback MPC}
In order to allow force feedback from sensors, we follow the approach described in~\cite{kleff:hal-04572399} which augments the position-velocity state with the contact force $\force$. The state and control are defined as:
\begin{equation}
\label{eq:state_control}
\begin{aligned}
    \state &= (\q,\qdot,\force) \in \R^{2n+m} \\
    \control &= \bm{\tau} \in \R^n 
\end{aligned}
\end{equation}

Contact forces are modeled with a linear spring–damper law as:
\begin{equation}
\label{eq:spring_damper}
\begin{aligned}
    \force(\q,\qdot) 
    &= -K \,\Delta p(\q,p_c) - B \,\dot p(\q,\qdot) .
\end{aligned}
\end{equation}
where $\Delta p(\q,p_c) = p(\q) - p_c$ is the end-effector deflection relative to the anchor $p_c \in \R^m$, $p(\q)\in\R^m$ is the forward kinematics, and $\dot p(\q,\qdot) = J(\q)\qdot$ is the Cartesian velocity. The matrices $K,B \in \R^{m\times m}$ are diagonal and positive definite.
Assuming no slip, differentiating~\eqref{eq:spring_damper} yields to the force dynamics:
\begin{equation}
\label{eq:force_dynamics}
\begin{aligned}
    \dot\force(\state,\control) 
    &= -K \,\dot p(\q,\qdot) - B \,\ddot p(\state,\control) \\
    \ddot p(\state,\control) 
    &= J(\q)\,\qddot(\q,\qdot,\force,\bm{\tau}) + \dot J(\q)\,\qdot
\end{aligned}
\end{equation}

The overall system dynamics combine joint and force dynamics as:
\begin{equation}
\label{eq:full_dynamics}
\begin{aligned}
    f_t &=
    \begin{bmatrix}
        \qdot \\[4pt]
        M^{-1}\!\left(\bm{\tau} - h + J^\top \force\right) \\[4pt]
        -KJ\qdot 
        - B\!\Big(JM^{-1}\!\left(\bm{\tau} - h + J^\top \force\right) 
        + \dot J \qdot\Big)
    \end{bmatrix}
\end{aligned}
\end{equation} 

This MPC formulation allows to combine the benefits of nonlinear MPC and direct force control and has been shown to outperform both taken individually in challenging contact tasks such as surface polishing and static multicontact pushing~\cite{kleff:hal-04572399}. However, tasks requiring the combined regulation of end-effector motion and tangential contact forces, such as deburring, have not been investigated yet. In particular, two contact models are considered for the deburring task: the $1D$ model which only predicts the normal contact force and assumes unconstrained tangential movement ($m=1$) and the $3D$ model which additionally predicts the tangential forces ($m=3$). 

\subsection{Cost and constraints}
The controller penalizes deviations in end–effector pose and contact force, while regularizing state and input effort. Hard constraints ensure feasibility under joint, torque, and collision limits. Throughout, the weighted norm is defined as $\|v\|_Q^2 := v^\top Q v$ for $Q \succeq 0$. 

\paragraph*{Stage and terminal costs}
The running and terminal costs are
\begin{equation}
\label{eq:costs}
\begin{aligned}
    \ell_t &=
        \|e_t\|_{Q_e}^2
      + \|\state_t - \state_t^{\mathrm{ref}}\|_{Q_x}^2
      + \|\control_t - \bm{\tau}(\q_t,\force_t)\|_{R_u}^2 \\
    \ell_T &=
        \|e_T\|_{Q_{e,T}}^2
      + \|\state_T - \state_T^{\mathrm{ref}}\|_{Q_{x,T}}^2
\end{aligned}
\end{equation}
where $Q_e = \mathrm{diag}(Q_p, Q_R, Q_{\force,\parallel}, Q_{\force,\perp})$ (and similarly for $Q_{e,T}$) is block diagonal to weight the residuals independently. The residual vector is defined as:

\begin{equation}
\label{eq:residual_components}
\begin{aligned}
    e_t &= \big[e_p(t), e_R(t), e_{\force,\parallel}(t), e_{\force,\perp}(t)\big]^\top \\
    e_p(t) &= p(\q_t) - p_t^{\mathrm{ref}} \in \R^3 \\
    e_R(t) &= \operatorname{vec}\!\left(\log\!\big(R_t^{\mathrm{ref}\,\top} R(\q_t)\big)\right) \in \R^3 \\
    e_{\force,\parallel}(t) &= \force_{t,\parallel} - \force_{t,\parallel}^{\mathrm{ref}} \in \R \\
    e_{\force,\perp}(t) &= \force_{t,\perp} \in \R^{2} ~~~~~\text{(3D model only)}
\end{aligned}
\end{equation}
\noindent where $p(\q)\in\R^3$ and $R(\q)\in \mathbb{SO}(3)$ are the forward kinematics of the end–effector position and orientation; $p_t^{\mathrm{ref}}$ and $R_t^{\mathrm{ref}}$ are their time-varying references; $\operatorname{vec}\!\big(\log(\cdot)\big) :\mathbb{SO}(3)\to\R^3$ denotes the orientation error as a rotation vector; $\force_t\in\R^m$ is the contact force; $n\in\R^3$ is the unit vector along the hole axis and $\force_{t,\parallel}^{\mathrm{ref}}\in\R$ is the desired normal contact force. The term $e_{\force,\parallel}$ enforces tracking of the desired normal force, while $e_{\force,\perp}$ penalizes tangential forces, which are defined as:
\begin{equation}
\label{eq:force_decomp}
\begin{aligned}
    \force_{t,\parallel} &= n^\top \force_t \in \R , \\
    \force_{t,\perp} &= (I - nn^\top)\force_t \in \R^{2} ~~~~~ \text{(3D model only)} .
\end{aligned}
\end{equation}
Note that the tangential force $\force_{t,\perp}$ and residual $e_{\force,\perp}(t)$ are only used with the $3D$ model ($m=3$). The reference $\bm{\tau}(\q_t,\force_t)$ is the gravity/Coriolis compensation torque, typically $\bm{g}(\q_t) - J(\q_t)^\top \force_t$, assuming zero reference velocity. It was tracked thanks to a regularization term $R_u$.

\paragraph*{Task-phase weighting}
The cost terms are scheduled across phases of the task:
\begin{enumerate}
    \item \textit{Trajectory following:} follow joint trajectory ($Q_x$ active), no force tracking ($Q_{\force,\parallel}=Q_{\force,\perp}=0$), no pose tracking ($Q_p=Q_R=0$).
    \item \textit{Pre-contact:} increase pose tracking ($Q_p,Q_R$), reduce reliance on reference configuration.
    \item \textit{Insertion:} increase $Q_{\force,\parallel}$ to build normal force, keep $Q_{\force,\perp}$ large to suppress tangential force, maintain orientation with $Q_R$, and disable reference configuration tracking ($Q_q=0$).
    \item \textit{Deburring:} track circular $p_t^{\mathrm{ref}}$ in the plane orthogonal to $n$, enforce normal force with $Q_{\force,\parallel}$, suppress tangential forces with $Q_{\force,\perp}$, and maintain tool orientation with $Q_R$.
\end{enumerate}

\paragraph*{Collision avoidance}
For each pair $(i,j)$ of robot–environment bodies, a signed-distance margin $\varepsilon>0$ is imposed:
\begin{equation}
\label{eq:collision_constraint}
    c_{t,i,j}(\q_t) := d_{ij}(\q_t) - \varepsilon \;\ge\; 0 ,
\end{equation}
where $d_{ij}(\q_t)$ is the signed distance between body $i$ and body $j$. For the trajectory optimization algorithm, based on Sequential-Quadratic Programming (SQP)~\cite{jordana:hal-04330251}, the following linearization is used~\cite{haffemayer:hal-04425002}: 
\begin{equation}
\label{eq:collision_linearization}
    c_{t,i,j}(\q_t) \approx c_{t,i,j}(\bar\q_t)
    + \nabla_{\!q} d_{ij}(\bar\q_t)^\top (\q_t-\bar\q_t) \;\ge\; 0
\end{equation}

\section{Diffusion priors as memory of motion} 
\label{sec:diffusion}
Collision avoidance in cluttered, non-strictly convex workspaces is hard for local optimizers, since the feasible set is highly non-convex and narrow passages are common. MPC  with generic initialization often converges to poor local minima, or spends many iterations just to recover feasibility. This motivates a warm-start that already encodes task-specific strategies and collision-aware geometry.
\subsection{Trajectory generation}
To do so, we build a synthetic dataset of feasible deburring motions that captures the multimodality of motions around the different holes and poses. The motion data are obtained by the task-and-motion planner Humanoid Path Planner \cite{mirabel2016hpp}, by defining a simple task sequence composed of 3 steps: reach a pre-grasp location in front of the hole, apply the contact, retract to a rest location. 

For each configuration we (i) solve inverse kinematics to reach the target end effector pose from many seeds, (ii) connect start and goal with a bi-directional RRT$^\star$ in configuration space to get a collision free path, (iii) refine this path with an OCP to obtain feasible and smooth trajectories. Since the dataset is inherently multimodal, with multiple approach directions and collision-avoidance strategies, we train a conditional diffusion motion to capture this diversity without mode collapse and enable controlled sampling, providing a reusable prior that encodes motion strategies.

\subsection{Diffusion model for memory of motion}
The model maps a noise sequence to a trajectory proposal conditioned on the initial robot state $\state_0$ and the target end-effector pose. We implement this with a Diffusion Transformer (DiT) \cite{peebles_scalable_2023}, following the standard diffusion architecture for trajectories \cite{ho_denoising_2020,janner_planning_2022}, where trajectory tokens represent joint configurations, while conditioning tokens encode $\state_0$ and the target pose using learnable type and timestep embeddings. An encoder builds a memory from the conditioning tokens, a decoder cross-attends to this memory while processing trajectory tokens, and a linear head predicts the noise across all timesteps. To satisfy real-time constraints, we restrict the diffusion process to a small number of denoising steps $D_s=25$, rather than the hundreds typically used in generative modeling, which keeps inference fast while preserving robustness and diversity, especially when combined with classifier-free conditioning.

\subsection{Using the diffusion as prior}
The diffusion output is not executed directly. Instead, the diffusion output serves as a reference for the OCP inside the MPC loop, similar to the approach in~\cite{haffemayer_collision-free_2025}. We use the state regularization term from~(\ref{eq:costs}), but set its reference to the diffusion prior trajectory. This effectively pulls the solution toward the prior while dynamics, force objectives, and hard constraints enforce feasibility. In practice, this gives the solver a favorable basin of attraction and accelerates convergence \cite{mansard_using_2018}, while MPC ensures force tracking, torque limits, and collision avoidance.

\section{Experiments}

\begin{figure}[!t]
    \centering
    \setlength{\belowcaptionskip}{-15pt}
   \includegraphics[width=0.45\linewidth]{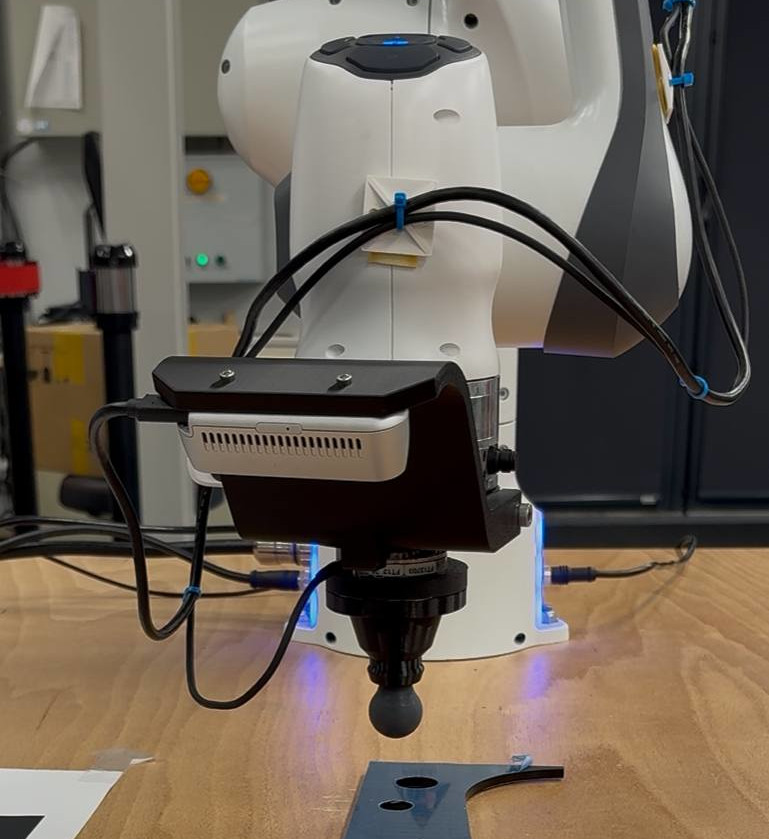}
       \vspace{0pt}
        \caption{Custom end-effector mount for the Franka Emika Panda integrating a force-torque sensor and provisions for RGB-Depth camera. A spherical tool tip enables smooth circular motion within holes. (Scenario 2, Sec. \ref{sec:scenario2})}
        \label{fig:custom-end-effector}
\end{figure}

\subsection{Setup}

A custom instrumented end-effector is attached to a Franka Emika Panda robot, as shown in Fig. \ref{fig:custom-end-effector}. The end effector is provided with an ATI Mini45 force-torque sensor and an Intel RealSense D435i RGB-D camera. The tool has a spherical shape to allow smooth circular deburring-like motions.   The Diffusion model is running on a computer with AMD Ryzen~9 5950X CPU with NVIDIA RTX 3060 GPU. The MPC is implemented in Python in Crocoddyl \cite{DBLP:journals/corr/abs-1909-04947} with CSQP solver \cite{jordana:hal-04330251} running inside of a real-time process of ros2\_control on a computer with Intel i9-14900K @5.6 GHz. Both computers are connected in the same network and communicate through ROS2 Humble.

Similarly to~\cite{kleff:hal-04572399}, the controlled torques are linearly interpolated at 1~kHz, and the measured forces are low-pass filtered using a second-order Butterworth filter (cut-off frequency: 10~Hz).

\subsection{Task definition}
\begin{figure}[!t]
    \centering
    \setlength{\abovecaptionskip}{-2pt}
   \includegraphics[width=0.95\linewidth]{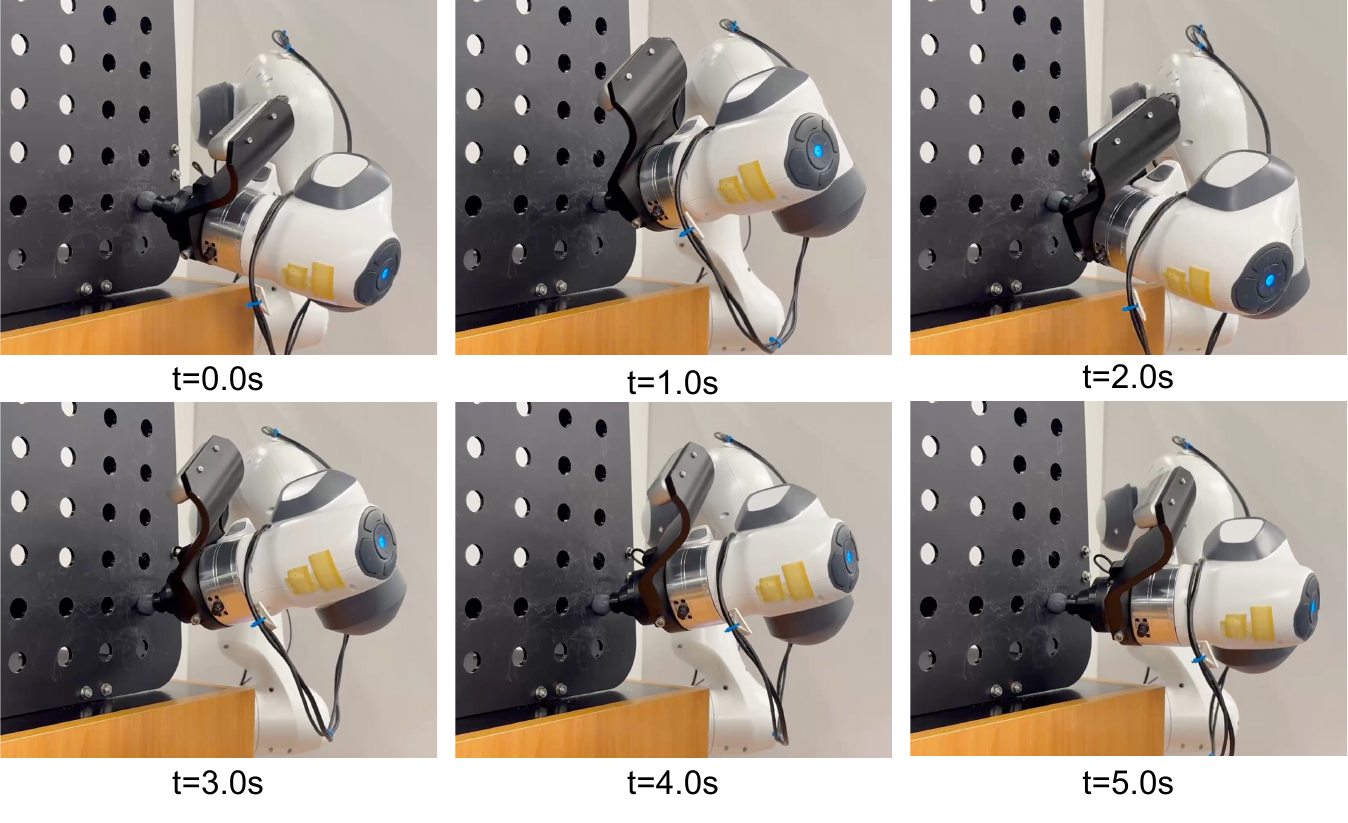}
        \caption{Representative sequence of a circular deburring motion performed onto the object. (Scenario 2, Sec. \ref{sec:scenario2})}
        \label{fig:deburring_motion}
\end{figure}

The robot is set to reach one or more holes with progressively increasing difficulty (Scenarios 1–3 described below). Upon reaching a hole, the robot regulates a normal deburring force of 30 N and executed a circular (revolute) passes at angles of 7.5~deg and 15~deg about the hole axis, each lasting 5~s. A typical deburring motion can be seen in Fig. \ref{fig:deburring_motion}. When activated, the collision constraint is imposed between a capsule approximation of the forearm and a planar obstacle in front of the robot.

\subsection{Scenario 1: Accessible Single-Hole Deburring} \label{sec:scenario1}


\begin{figure}[!t]
    \centering
   \includegraphics[width=1.0\linewidth]{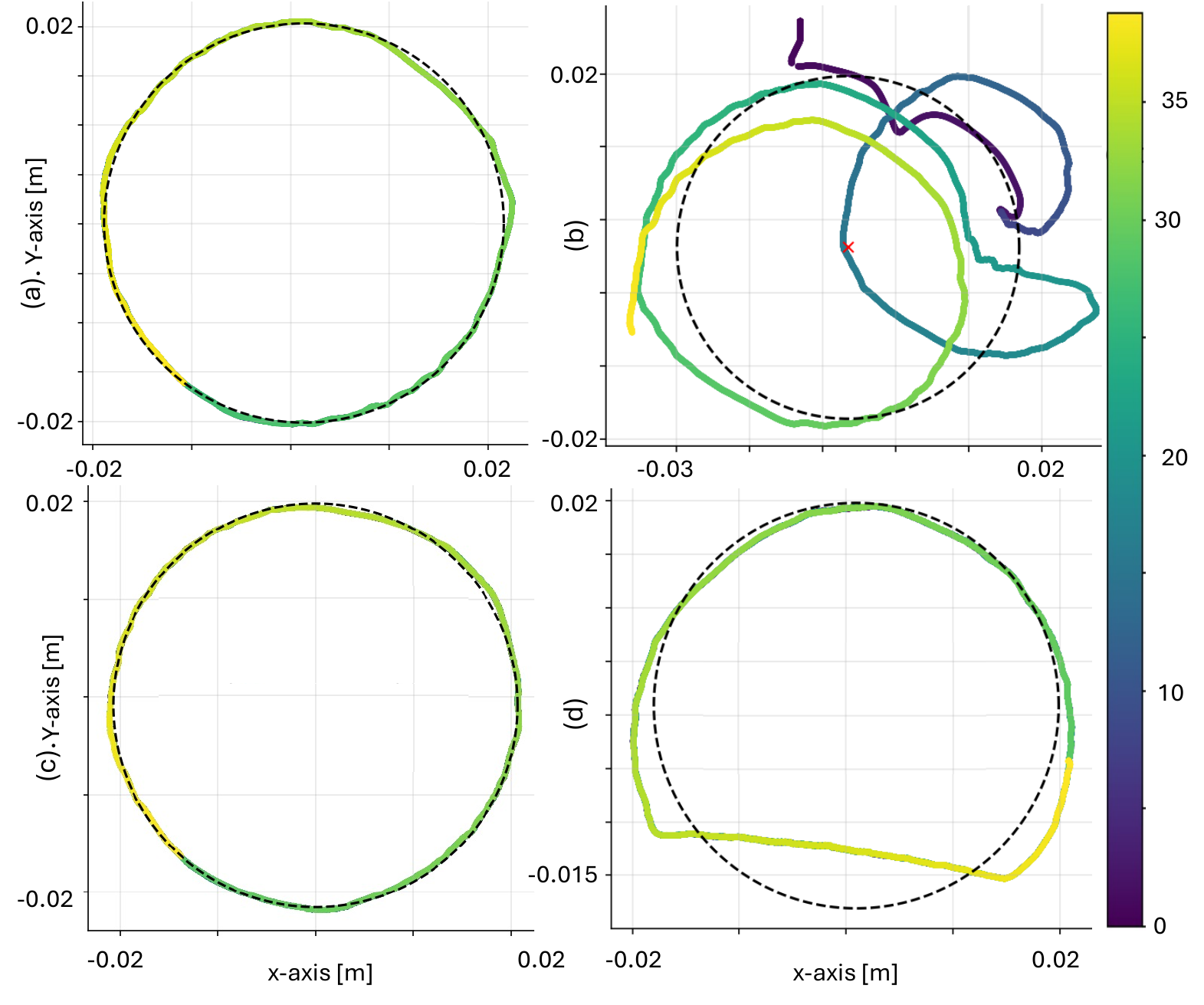}
      \setlength{\belowcaptionskip}{-20pt}
      \setlength{\abovecaptionskip}{-10pt}
       \vspace{0pt}
               \caption{Scenario 1 (Sec. \ref{sec:scenario1}): Force-colored end-effector x-y trajectories for a single hole; color encodes normal normal force magnitude. (a) 1D contact, no obstacle avoidance; (b) 1D contact with an obstacle in front of the robot; (c) 3D contact, no collision avoidance; (d) 3D contact with collision avoidance. The obstacle shape is clearly visible in the successful example c, and prevents the robot to achieve the complete circle.} 

        \label{fig:scenario1_forces}
\end{figure}

\begin{figure}[!t]
    \centering
   \includegraphics[width=0.9\linewidth]{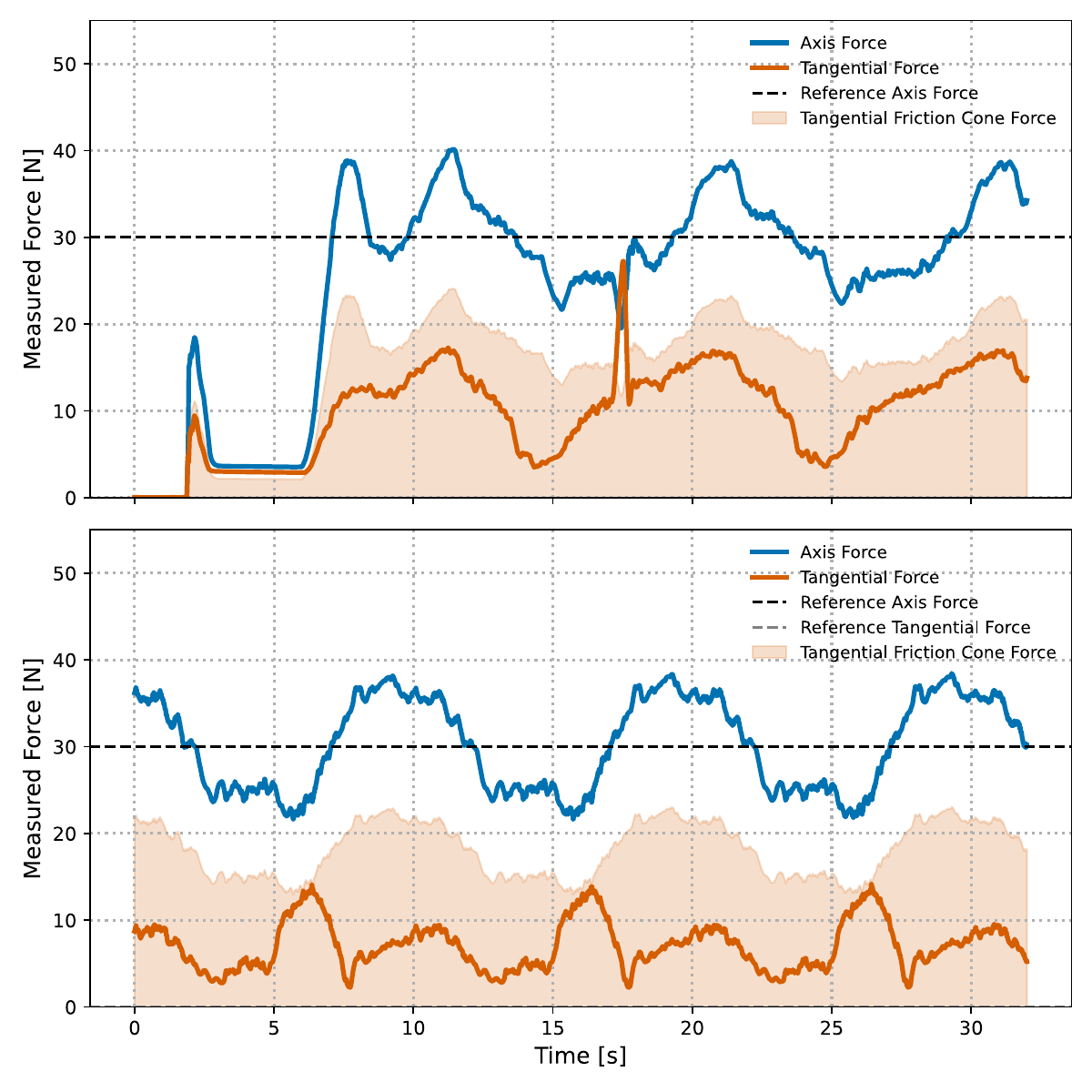}
      \setlength{\belowcaptionskip}{-20pt}
      \setlength{\abovecaptionskip}{-2pt}
         \caption{Scenario 1 (Sec. \ref{sec:scenario1}): Tracking of normal and tangential forces during deburring motion on a flat table with collision avoidance with 1D (top) and 3D contact (bottom) models. In this case when 1D contact model approaches collision, lateral forces exceed force friction cone ($\mu = 0.6$) and break the contact while 3D contact maintains stable movement.}  
        \label{fig:scenario1_forces_timeseries}
\end{figure}

\begin{figure*}[t] 
  \centering
  \setlength{\belowcaptionskip}{-10pt}
  \includegraphics[width=0.77\textwidth]{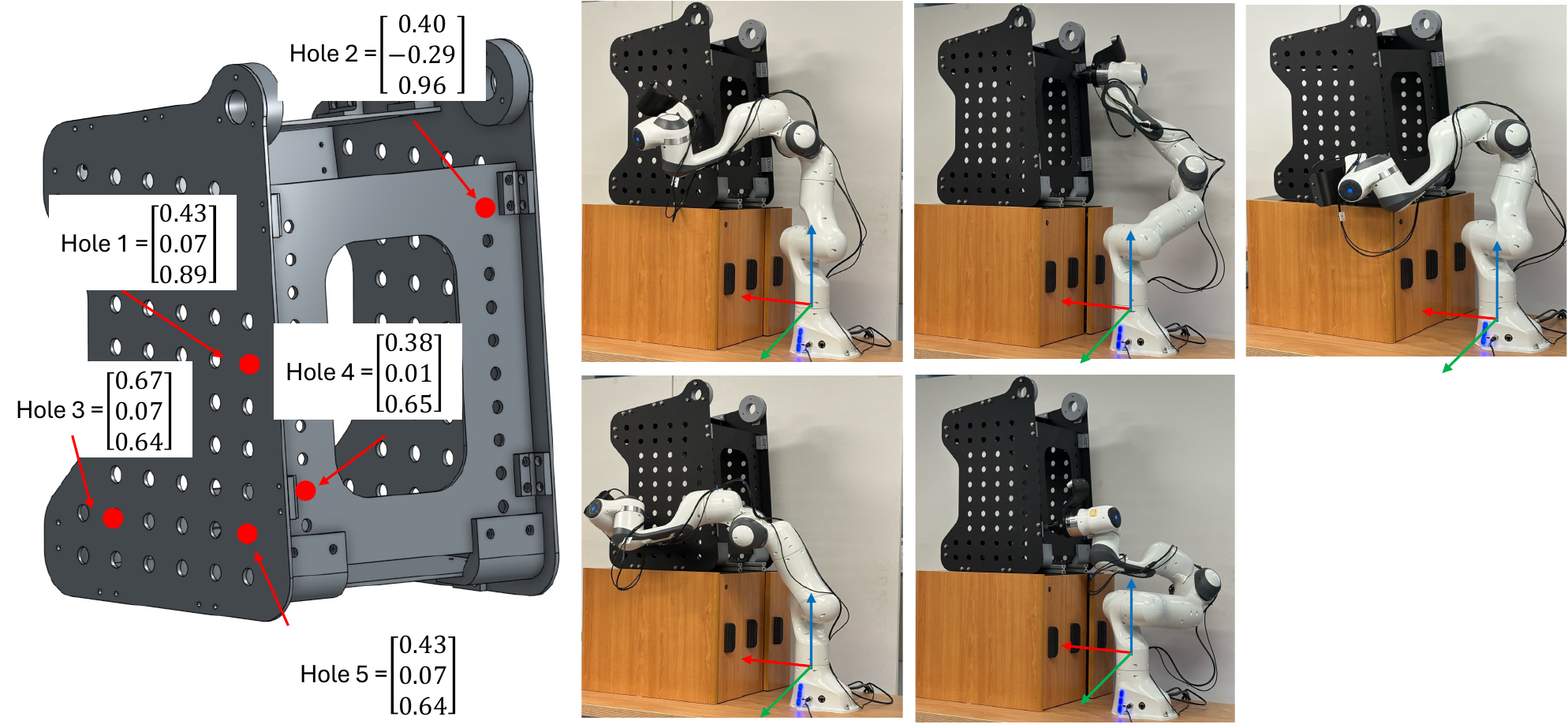}
  \caption{Positions of the 5 holes to be deburred  on the investigated workpiece and reaching postures for each hole.  }
  \label{fig:scenario_3}
\end{figure*}

This experiment serves as the method baseline for the force feedback MPC. 
It also evaluates 1D and 3D contact models under hard collision avoidance constraints. 
As explained earlier, the 1D model regulates only the normal force along the tool axis, whereas the 3D model also accounts for tangential forces. 
The setup, visible in Fig.  \ref{fig:custom-end-effector}, consists of a very accessible single hole positioned on a table in front of the robot ($x = 0.52$ m, $y = 0.0$ m, in robot base frame shown in Fig. \ref{fig:scenario_3}) at the same height as the robot's base. Identical cost weights and gains were used across conditions to ensure fair comparison. For these experiments, the MPC was executed at 200~Hz with a 300~ms horizon and 15 nodes, using 3~SQP iterations and 150~QP iterations per cycle. 
In the nominal setting, the robot performs the deburring task without any obstacle. As shown in Fig. \ref{fig:scenario1_forces}a and \ref{fig:scenario1_forces}c, when there is no collision constraint, both models were able to correctly track desired end-effector trajectories. The normal force tracking for both models in this obstacle-free setting along four repetitions of the deburring task is shown in Fig. \ref{fig:scenario1_forces_timeseries}a and Fig. \ref{fig:scenario1_forces_timeseries}b. The Root Mean Square Error (RMSE) was $5.35$ N and $4.40$ N, for 1D and 3D models respectively.
In the nominal case both models perform equally well.

A second setup placed the robot close to a wall (i.e. close to a collision). In this setting, the nominal circular deburring motion would make the robot collide with the wall. 
For the 3D model, activating the collision-avoidance constraint in \eqref{eq:collision_constraint} forced the solver trade-off trajectory tracking against constraint satisfaction and deviate the end-effector path during execution (Fig.~\ref{fig:scenario1_forces}d). 
Under these conditions, the 1D contact model failed to maintain stable in-hole contact (Fig.~\ref{fig:scenario1_forces}b). The tool slid out of the hole as the optimizer prioritized clearance over orientation, while tangential forces were not explicitly regulated. This led to violation of the no-slip condition. This setup confirms that the explicit regulation of tangential forces is crucial to preserve contact stability.





\subsection{Scenario 2: Obstacle-Constrained Multi-Hole Reach-and-Deburring Sequence} \label{sec:scenario2}

\begin{figure}[!t]
    \centering
    \setlength{\belowcaptionskip}{-20pt}
   \includegraphics[trim={0 2.5cm 0 2.5cm},clip,width=1.0\linewidth]{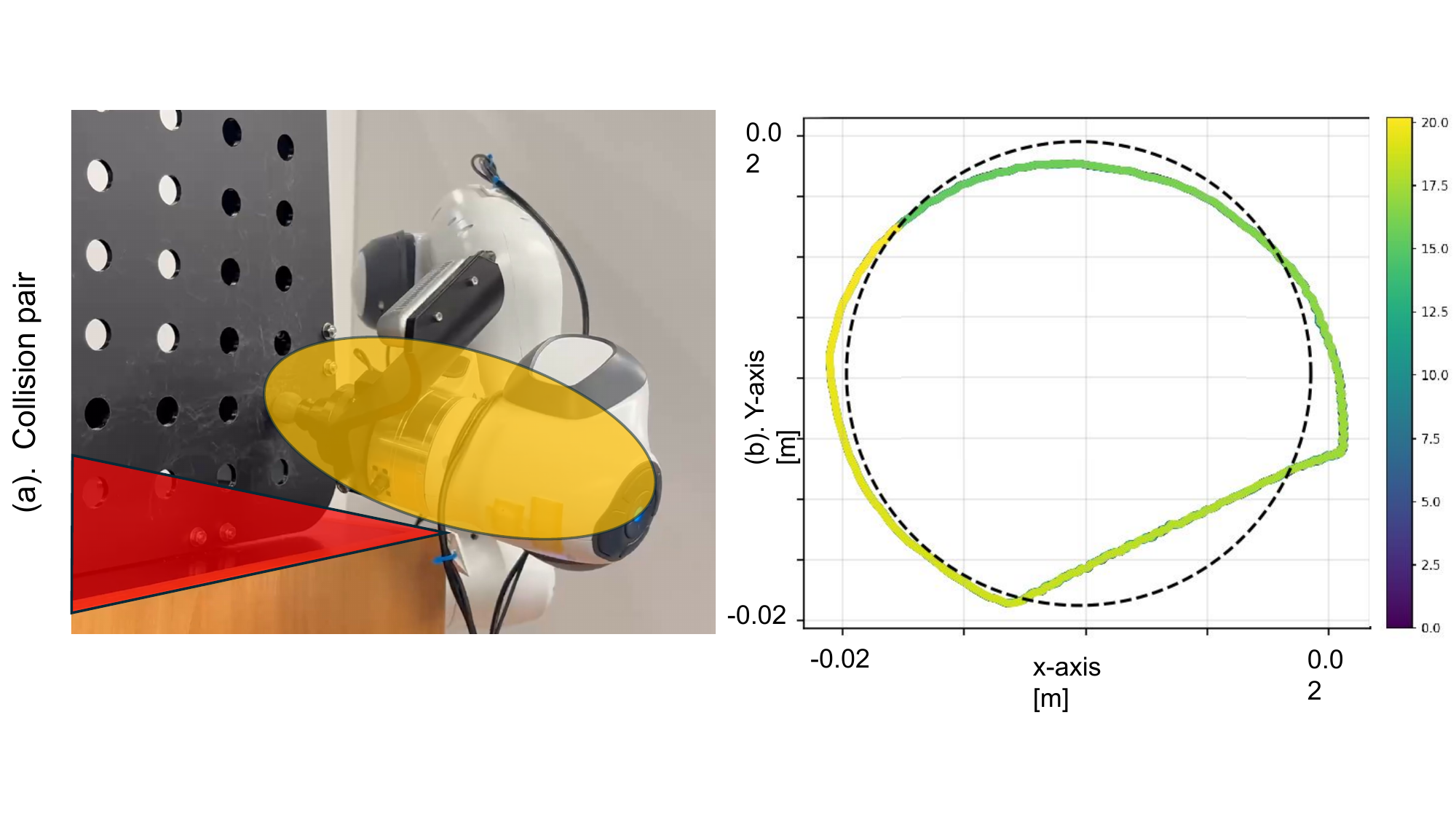}
        \caption{Description of the collision pair between the plane (red) and ellipsoid (yellow) (a). Force-colored  end-effector x-y trajectories  for a single hole. Color encodes normal force magnitude.}  
        \label{fig:scenario_3_forces}
\end{figure}

This scenario evaluates force–tracking performance in more demanding configurations involving both collision-avoidance constraints and far-reach targets (Fig.~\ref{fig:scenario_3}a). The workpiece contains holes intentionally placed near the robot’s workspace boundary: the robot’s maximum nominal reach is 0.85~m, whereas Holes~3 and~2 are located at radial distances of  0.92~m and 1.00~m, respectively. Additional holes are positioned to highlight the effect of a plane collision constraint defined at the base of the workpiece (Fig.~\ref{fig:scenario_3_forces}a). The resulting five reaching postures are shown in Fig.~\ref{fig:scenario_3}; as expected, holes~3–2 induce near fully-extended arm configurations. Thus, as shown in Table \ref{tab:inv_manip_rmse} this induces a reduced force manipulability. \footnote{Force manipulability \cite{yoshikawa1985manipulability} is calculated as $w_{\text{f}}=\frac{1}{\sqrt(det(JJ^T))}$.} To improve stability in these configurations, the controller is executed at 500~Hz with a prediction horizon of 30~ms, discretized into 15 nodes. The solver uses 3~SQP iterations and 25~QP iterations per cycle. No collision constraints are applied in this task.

\begin{table}[t]
  \centering
  \caption{Inverse velocity manipulability and corresponding force RMSE.}
  \label{tab:inv_manip_rmse}
  \begin{tabular}{lcc}
    \toprule
    Hole & $w_{\text{f}}$ (--) & Force RMSE [N] \\
    \midrule
    Hole 1 & 10.6956 & 11.30 \\
    Hole 2 &  6.4821 & 7.98 \\
    Hole 3 &  6.9166 & 17.20 \\
    Hole 4 & 19.0211 & 17.41 \\
    Hole 5 & 12.3884 & 15.34 \\
    \bottomrule
  \end{tabular}
  \vspace{-15pt}
\end{table}

Across all five targets, deburring is completed without slip or collisions (success rate of 100\%), indicating that the controller can jointly satisfy contact and geometric constraints even in adverse poses. However, the normal-force tracking RMSE increases as the robot operates farther from the baseline pose of Scenario~1 (Fig.~\ref{fig:scenario_3}c). This degradation is consistent with (i) poor Jacobian conditioning near workspace limits, which reduces force authority along the hole axis; (ii) proximity to joint/torque bounds that constrains corrective actions; and (iii) tighter orientation feasibility when the end-effector must remain aligned with the hole while respecting the plane constraint. In practice, these factors reduce the effective bandwidth available for force regulation and increase sensitivity to small pose deviations or modeling errors.
For holes adjacent to the plane constraint, the nominal circular deburring trajectory can not be executed in full, as the avoidance constraint truncated the path, which is an outcome analogous to Scenario~1 but amplified by the far-reach geometry. As shown in Fig.~\ref{fig:scenario_3_forces}. In these cases the solver produces collision-aware path deflections that preserves contact stability while trading off perfect trajectory tracking against the maintenance of the desired normal force.



\subsection{Scenario 3: Long Sequence Integration}

In this scenario visible in the accompanying video, it is shown a complete integration of the deburring task on the MPC level with a diffusion-guided motion around the object. Here we demonstrate the complete integration of the pipeline. An ad-hoc sequence decides the list of holes to be machined. For each hole, the diffusion model is queried to produce a reference guide used as specified in Sec. \ref{sec:diffusion}.

The MPC executes the motion while regulating the contact force. It compensates for model inaccuracies and object pose deviations, while simultaneously enforcing local collision avoidance. The diffusion guidance allows the robot to reach difficult locations beyond narrow passages which would be very expensive to compute at run-time with an on-line planner. Holes are reached and machined within the accuracy range observed in Scenario \ref{sec:scenario2}. As shown in the video, the torque controlled compliance also tolerates physical disturbance by a human co-operator.

\section{Discussions}



The proposed method enabled us to perform a complex industrial task by integrating capabilities that are often treated separately in robotics workflows:  collision avoidance, reactive control, accurate force feedback. While we have argued that the approach is broadly applicable, its current performance relies on several assumptions and design choices that warrant discussion.

First, our approach relies on offline computations using a motion planner to train a diffusion model. In this work, we considered static scenes, so the computation cost and sample requirements are not representative of what would be needed to generalize to unknown environments, ideally by conditioning on sensor-based representations of obstacles and targets. Future work will explore parallelized motion planning on GPU \cite{bialkowski2011massively}, which could be integrated into our current workflow with minimal modifications. Moreover, exhaustive offline exploration could be alleviated by enabling fast refinement of the motion prior at run time, for instance with sample-based optimizers such as MPPI \cite{huang2025prrtc}, which naturally provide strong exploration capabilities.

At high control frequencies, MPC cannot efficiently handle a large number of collision pairs. To reach $200~Hz$ in our experiments, we indeed restricted the controller to a single pair. This limitation is largely mitigated by the motion prior, which provides safe, collision-free guidance, but it may become restrictive in more complex scenarios. Achieving such performance also requires careful tuning of the numerical solver (e.g., QP iterations), a know-how that could be alleviated by adopting recent advances in numerical optimization \cite{frison2020hpipm}.

The experiments also highlight that a key limiting factor lies in the robot kinematics and link lengths. In such complex scenes, the robot struggles to reach all the holes shown in Fig.~\ref{fig:panda} while simultaneously maintaining the prescribed circular motion. Our results show that task accuracy is strongly correlated with manipulability. More generally, far-reach and obstacle-constrained conditions stress both motion planning and contact regulation, motivating future work on manipulability-aware cost shaping \cite{maric2019fast, kennel2021manipulability} and constraint handling to preserve force fidelity near workspace limits. One possible direction is to introduce a cost term such as $\log(\det(J^T))$ to bias reaching postures toward configurations with higher directional force authority. However, the nonlinearity of such a term may negatively impact solver performance, which suggests investigating local convexifications or scheduling strategies to maintain a tractable QP at each MPC step.

It would also be beneficial to transfer the task to a mobile manipulator. The added mobility introduces the need for a careful sequencing of the holes to be machined, to minimize unnecessary base movements. External perception, e.g. through vision-based system, would become essential to coordinate the robot’s repositioning and ensure accuracy.



\section{Conclusion}



We presented a complete solution based on model predictive control to perform an industrial-inspired deburring task. A controller enables the robot to safely navigate around complex obstacles, avoid collisions despite disturbances, and maintain precise force regulation through direct sensor feedback. Operating in torque control further enhances safety: the robot can be interrupted by a nearby worker, and the risk of damage is reduced even in the presence of undetected obstacles, making the approach suitable for human–robot collaboration.

Our framework builds on open-source modules for motion planning, optimization, and ROS integration, and we will release the code upon acceptance. By combining these elements, the method provides a practical alternative to typical robot deburring setups, which rely on costly tools, rigid manipulators, and dedicated insertion guides. Instead, we show that a collaborative robot can achieve comparable performance and strategies to a human operator, while remaining versatile enough to be deployed for a wide range of collaborative tasks using standard tools and processes. Future work will focus on extending the proposed methods beyond the current limitations and transferring the demonstrator to more realistic scenarios involving direct metal–tool interaction.






\bibliographystyle{IEEEtran}
\bibliography{IEEEabrv,bibliography}

@unpublished{kleff:hal-04572399,
  TITLE = {{Force Feedback in Model Predictive Control: A Soft Contact Approach}},
  AUTHOR = {Kleff, S{\'e}bastien and Jordana, Armand and Mansard, Nicolas and Righetti, Ludovic},
  URL = {https://hal.science/hal-04572399},
  NOTE = {working paper or preprint},
  HAL_LOCAL_REFERENCE = {Rapport LAAS n{\textdegree} 24093},
  YEAR = {2024},
  MONTH = May,
  HAL_ID = {hal-04572399},
  HAL_VERSION = {v1},
}

@INPROCEEDINGS{Farshidian2017,
  author={Farshidian, Farbod and Jelavic, Edo and Satapathy, Asutosh and Giftthaler, Markus and Buchli, Jonas},
  booktitle={IEEE-RAS International Conference on Humanoid Robotics (Humanoids)}, 
  title={Real-time motion planning of legged robots: A model predictive control approach}, 
  year={2017},
  volume={},
  number={},
  doi={10.1109/HUMANOIDS.2017.8246930}}

@article{Mastalli2023,
archivePrefix = {arXiv},
arxivId = {2209.05375},
author = {Mastalli, Carlos and Chhatoi, Saroj Prasad and Corberes, Thomas and Tonneau, Steve and Vijayakumar, Sethu},
doi = {10.1109/TRO.2023.3262186},
eprint = {2209.05375},
issn = {19410468},
journal = {IEEE Transactions on Robotics},
number = {4},
pages = {3222--3241},
title = {{Inverse-Dynamics MPC via Nullspace Resolution}},
volume = {39},
year = {2023}
}

@article{Mason1981,
author = {Mason, Matthew T.},
doi = {10.1109/TSMC.1981.4308708},
issn = {21682909},
journal = {IEEE Transactions on Systems, Man and Cybernetics},
number = {6},
pages = {418--432},
title = {{Compliance and Force Control for Computer Controlled Manipulators}},
volume = {11},
year = {1981}
}

@article{Matschek2017,
author = {Matschek, Janine and Bethge, Johanna and Zometa, Pablo and Findeisen, Rolf},
doi = {10.1016/J.IFACOL.2017.08.898},
issn = {2405-8963},
journal = {IFAC-PapersOnLine},
publisher = {Elsevier},
title = {{Force Feedback and Path Following using Predictive Control: Concept and Application to a Lightweight Robot}},
volume = {50},
year = {2017}
}

@article{Wahrburg2016,
author = {Wahrburg, Arne and Listmann, Kim},
doi = {10.1109/CDC.2016.7799435},
isbn = {9781509018376},
journal = {2016 IEEE 55th Conference on Decision and Control, CDC 2016},
month = {dec},
pages = {7548--7554},
publisher = {Institute of Electrical and Electronics Engineers Inc.},
title = {{MPC-based admittance control for robotic manipulators}},
year = {2016}
}

@ARTICLE{Gazar21,
  author={Gazar, Ahmad and Nava, Gabriele and Chavez, Francisco Javier Andrade and Pucci, Daniele},
  journal={IEEE Transactions on Robotics}, 
  title={Jerk Control of Floating Base Systems With Contact-Stable Parameterized Force Feedback}, 
  year={2021},
  volume={37},
  number={1},
  pages={1-15},
  doi={10.1109/TRO.2020.3005547}}

@article{Hogan1985,
author = {Hogan, Neville},
journal = {J. Dyn. Sys., Meas., Control.},
pages = {1-24},
title = {{Impedance Control Part1-3}},
volume = {107},
year = {1985}
}

@article{Neunert2018,
archivePrefix = {arXiv},
arxivId = {1712.02889},
author = {Neunert, Michael and Stauble, Markus and Giftthaler, Markus and Bellicoso, Carmine Dario and Carius, Jan and Gehring, Christian and Hutter, Marco and Buchli, Jonas},
doi = {10.1109/LRA.2018.2800124},
eprint = {1712.02889},
issn = {23773766},
journal = {IEEE Robotics and Automation Letters},
number = {3},
pages = {1458--1465},
title = {{Whole-Body Nonlinear Model Predictive Control Through Contacts for Quadrupeds}},
volume = {3},
year = {2018}
}

@INPROCEEDINGS{Kleff22,
  author={Kleff, Sébastien and Dantec, Ewen and Saurel, Guilhem and Mansard, Nicolas and Righetti, Ludovic},
  booktitle={2022 IEEE/RSJ International Conference on Intelligent Robots and Systems (IROS)}, 
  title={Introducing Force Feedback in Model Predictive Control}, 
  year={2022},
  volume={},
  number={},
  pages={13379-13385},
  doi={10.1109/IROS47612.2022.9982003}}

@article{Fahmi2020,
archivePrefix = {arXiv},
arxivId = {1904.12306},
author = {Fahmi, Shamel and Focchi, Michele and Radulescu, Andreea and Fink, Geoff and Barasuol, Victor and Semini, Claudio},
doi = {10.1109/TRO.2019.2954670},
eprint = {1904.12306},
issn = {19410468},
journal = {IEEE TRO},
month = {apr},
number = {2},
pages = {},
publisher = {Institute of Electrical and Electronics Engineers Inc.},
title = {{STANCE: Locomotion Adaptation over Soft Terrain}},
volume = {36},
year = {2020}
}

@ARTICLE{Selvaggio2023,
  author={Selvaggio, Mario and Garg, Akash and Ruggiero, Fabio and Oriolo, Giuseppe and Siciliano, Bruno},
  journal={IEEE Transactions on Control Systems Technology}, 
  title={Non-Prehensile Object Transportation via Model Predictive Non-Sliding Manipulation Control}, 
  year={2023},
  volume={31},
  number={5},
  pages={2231-2244},
  doi={10.1109/TCST.2023.3277224}}

@INPROCEEDINGS{Muller2019,
  author={Müller, Daniel and Mayer, Annika and Sawodny, Oliver},
  booktitle={2019 American Control Conference (ACC)}, 
  title={Model Predictive Force Control for Robots in compliant Environments with guaranteed Maximum Force}, 
  year={2019},
  volume={},
  number={},
  pages={1355-1360},
  doi={10.23919/ACC.2019.8814664}}

@article{Pankert2020,
author = {Pankert, Johannes and Hutter, Marco},
doi = {10.1109/LRA.2020.3010721},
issn = {23773766},
journal = {IEEE Robotics and Automation Letters},
month = {oct},
number = {4},
pages = {6177--6184},
publisher = {Institute of Electrical and Electronics Engineers Inc.},
title = {{Perceptive model predictive control for continuous mobile manipulation}},
volume = {5},
year = {2020}
}

@article{Kazim2018,
author = {Kazim, Khalid J. and Bethge, Johanna and Matschek, Janine and Findeisen, Rolf},
doi = {10.23919/ACC.2018.8431272},
isbn = {9781538654286},
issn = {07431619},
journal = {Proceedings of the American Control Conference},
month = {aug},
pages = {3153--3158},
publisher = {Institute of Electrical and Electronics Engineers Inc.},
title = {{Combined Predictive Path Following and Admittance Control}},
volume = {2018-June},
year = {2018}
}

@article{Minniti2021,
archivePrefix = {arXiv},
arxivId = {2106.04202},
author = {Minniti, Maria Vittoria and Grandia, Ruben and F{\"{a}}h, Kevin and Farshidian, Farbod and Hutter, Marco},
doi = {10.1109/ICRA48506.2021.9562066},
eprint = {2106.04202},
isbn = {9781728190778},
issn = {10504729},
journal = {Proceedings - IEEE International Conference on Robotics and Automation},
number = {Icra},
pages = {1651--1657},
title = {{Model Predictive Robot-Environment Interaction Control for Mobile Manipulation Tasks}},
volume = {2021-May},
year = {2021}
}

@article{Husmann2019,
author = {Husmann, S. and Stemmler, S. and H{\"{a}}hnel, S. and Vogelgesang, S. and Abel, D. and Bergs, T.},
doi = {10.1016/J.IFACOL.2019.11.459},
issn = {24058963},
journal = {IFAC-PapersOnLine},
publisher = {Elsevier B.V.},
title = {{Model Predictive Force Control in Grinding based on a Lightweight Robot}},
volume = {52},
year = {2019}
}

@article{Killpack2016,
author = {Killpack, Marc D. and Kapusta, Ariel and Kemp, Charles C.},
doi = {10.1007/s10514-015-9492-6},
issn = {15737527},
journal = {Autonomous Robots},
number = {3},
pages = {537--560},
publisher = {Springer US},
title = {{Model predictive control for fast reaching in clutter}},
volume = {40},
year = {2016}
}

@article{Grandia2023,
archivePrefix = {arXiv},
arxivId = {2208.08373},
author = {Grandia, Ruben and Jenelten, Fabian and Yang, Shaohui and Farshidian, Farbod and Hutter, Marco},
doi = {10.1109/TRO.2023.3275384},
eprint = {2208.08373},
issn = {19410468},
journal = {IEEE Transactions on Robotics},
number = {5},
pages = {3402--3421},
publisher = {IEEE},
title = {{Perceptive Locomotion Through Nonlinear Model-Predictive Control}},
volume = {39},
year = {2023}
}

@inproceedings{haffemayer:hal-04425002,
  TITLE = {{Model predictive control under hard collision avoidance constraints for a robotic arm}},
  AUTHOR = {Haffemayer, Arthur and Jordana, Armand and Fourmy, M{\'e}d{\'e}ric and Wojciechowski, Krzysztof and Saurel, Guilhem and Petr{\'i}k, Vladim{\'i}r and Lamiraux, Florent and Mansard, Nicolas},
  URL = {https://laas.hal.science/hal-04425002},
  BOOKTITLE = {{Ubiquitous Robots 2024}},
  ADDRESS = {New York (USA), United States},
  ORGANIZATION = {{Korea Robotics Society}},
  HAL_LOCAL_REFERENCE = {Rapport LAAS n{\textdegree} 24003},
  YEAR = {2024},
  MONTH = Jun,
  DOI = {10.1109/UR61395.2024.10597485},
  HAL_ID = {hal-04425002},
  HAL_VERSION = {v2},
}

@Inbook{Villani2008,
author="Villani, Luigi
and De Schutter, Joris",
editor="Siciliano, Bruno
and Khatib, Oussama",
title="Force Control",
bookTitle="Springer Handbook of Robotics",
year="2008",
publisher="Springer Berlin Heidelberg",
address="Berlin, Heidelberg",
pages="161--185",
isbn="978-3-540-30301-5",
doi="10.1007/978-3-540-30301-5_8",
url="https://doi.org/10.1007/978-3-540-30301-5_8"
}

@article{Whitney1977ForceFC,
  title={Force Feedback Control of Manipulator Fine Motions},
  author={Daniel E. Whitney},
  journal={Journal of Dynamic Systems Measurement and Control-transactions of The Asme},
  year={1977},
  volume={99},
  pages={91-97},
  url={https://api.semanticscholar.org/CorpusID:121383894}
}

@inproceedings{gaertner_collision-free_2021,
	title = {Collision-{Free} {MPC} for {Legged} {Robots} in {Static} and {Dynamic} {Scenes}},
	url = {https://ieeexplore.ieee.org/document/9561326},
	doi = {10.1109/ICRA48506.2021.9561326},
	urldate = {2025-08-28},
	booktitle = {2021 {IEEE} {International} {Conference} on {Robotics} and {Automation} ({ICRA})},
	author = {Gaertner, Magnus and Bjelonic, Marko and Farshidian, Farbod and Hutter, Marco},
	month = may,
	year = {2021},
	note = {ISSN: 2577-087X},
	pages = {8266--8272},
}

@inproceedings{haffemayer_collision_2025,
	title = {Collision {Avoidance} in {Model} {Predictive} {Control} using {Velocity} {Damper}},
	url = {https://laas.hal.science/hal-04707324},
	booktitle = {2025 {IEEE} {International} {Conference} on {Robotics} and {Automation} ({ICRA})},
	author = {Haffemayer, Arthur and Jordana, Armand and de Matteïs, Ludovic and Wojciechowski, Krzysztof and Righetti, Ludovic and Lamiraux, Florent and Mansard, Nicolas},
	month = may,
	year = {2025},
}

@article{DBLP:journals/corr/abs-1909-04947,
  author       = {Carlos Mastalli and
                  Rohan Budhiraja and
                  Wolfgang Merkt and
                  Guilhem Saurel and
                  Bilal Hammoud and
                  Maximilien Naveau and
                  Justin Carpentier and
                  Sethu Vijayakumar and
                  Nicolas Mansard},
  title        = {Crocoddyl: An Efficient and Versatile Framework for Multi-Contact
                  Optimal Control},
  journal      = {CoRR},
  volume       = {abs/1909.04947},
  year         = {2019},
  url          = {http://arxiv.org/abs/1909.04947},
  eprinttype    = {arXiv},
  eprint       = {1909.04947},
  bibsource    = {dblp computer science bibliography, https://dblp.org}
}

@misc{grandia_feedback_2019,
	title = {Feedback {MPC} for {Torque}-{Controlled} {Legged} {Robots}},
	url = {http://arxiv.org/abs/1905.06144},
	language = {en},
	urldate = {2022-10-24},
	publisher = {arXiv},
	author = {Grandia, Ruben and Farshidian, Farbod and Ranftl, René and Hutter, Marco},
	month = aug,
	year = {2019},
	note = {arXiv:1905.06144 [cs]},
}

@inproceedings{romualdi_online_2022,
	title = {Online {Non}-linear {Centroidal} {MPC} for {Humanoid} {Robot} {Locomotion} with {Step} {Adjustment}},
	doi = {10.1109/ICRA46639.2022.9811670},
	booktitle = {2022 {International} {Conference} on {Robotics} and {Automation} ({ICRA})},
	author = {Romualdi, Giulio and Dafarra, Stefano and L'Erario, Giuseppe and Sorrentino, Ines and Traversaro, Silvio and Pucci, Daniele},
	year = {2022},
	pages = {10412--10419},
}

@inproceedings{dantec_whole-body_2022,
	title = {Whole-{Body} {Model} {Predictive} {Control} for {Biped} {Locomotion} on a {Torque}-{Controlled} {Humanoid} {Robot}},
	doi = {10.1109/Humanoids53995.2022.10000129},
	booktitle = {2022 {IEEE}-{RAS} 21st {International} {Conference} on {Humanoid} {Robots} ({Humanoids})},
	author = {Dantec, Ewen and Naveau, Maximilien and Fernbach, Pierre and Villa, Nahuel and Saurel, Guilhem and Stasse, Olivier and Taix, Michel and Mansard, Nicolas},
	year = {2022},
	pages = {638--644},
}

@INPROCEEDINGS{Budhiraja2018,
  author={Budhiraja, Rohan and Carpentier, Justin and Mastalli, Carlos and Mansard, Nicolas},
  booktitle={IEEE Humanoids}, 
  title={Differential Dynamic Programming for Multi-Phase Rigid Contact Dynamics}, 
  year={2018},
  volume={},
  number={},
  doi={10.1109/HUMANOIDS.2018.8624925}}

@article{jordana:hal-04330251,
  TITLE = {{Structure-Exploiting Sequential Quadratic Programming for Model-Predictive Control}},
  AUTHOR = {Jordana, Armand and Kleff, S{\'e}bastien and Meduri, Avadesh and Carpentier, Justin and Mansard, Nicolas and Righetti, Ludovic},
  URL = {https://laas.hal.science/hal-04330251},
  JOURNAL = {{IEEE Transactions on Robotics}},
  HAL_LOCAL_REFERENCE = {Rapport LAAS n{\textdegree} 23425},
  PUBLISHER = {{IEEE}},
  YEAR = {2025},
  MONTH = Aug,
  HAL_ID = {hal-04330251},
  HAL_VERSION = {v3},
}

@article{gold_model_2023,
	title = {Model {Predictive} {Interaction} {Control} for {Robotic} {Manipulation} {Tasks}},
	volume = {39},
	doi = {10.1109/TRO.2022.3196607},
	number = {1},
	journal = {IEEE Transactions on Robotics},
	author = {Gold, Tobias and Völz, Andreas and Graichen, Knut},
	year = {2023},
	pages = {76--89},
}

@article{wijayarathne_real-time_2023,
	title = {Real-{Time} {Deformable}-{Contact}-{Aware} {Model} {Predictive} {Control} for {Force}-{Modulated} {Manipulation}},
	volume = {39},
	doi = {10.1109/TRO.2023.3286070},
	number = {5},
	journal = {IEEE Transactions on Robotics},
	author = {Wijayarathne, Lasitha and Zhou, Ziyi and Zhao, Ye and Hammond, Frank L.},
	year = {2023},
	pages = {3549--3566},
}

@misc{zhou_diffusion_2024,
	title = {Diffusion {Model} {Predictive} {Control}},
	url = {http://arxiv.org/abs/2410.05364},
	doi = {10.48550/arXiv.2410.05364},
	language = {en},
	urldate = {2025-05-16},
	publisher = {arXiv},
	author = {Zhou, Guangyao and Swaminathan, Sivaramakrishnan and Raju, Rajkumar Vasudeva and Guntupalli, J. Swaroop and Lehrach, Wolfgang and Ortiz, Joseph and Dedieu, Antoine and Lázaro-Gredilla, Miguel and Murphy, Kevin},
	month = oct,
	year = {2024},
	note = {arXiv:2410.05364 [cs]},
}

@article{chi_diffusion_2024,
	title = {Diffusion policy: {Visuomotor} policy learning via action diffusion},
	issn = {0278-3649},
	shorttitle = {Diffusion policy},
	url = {https://doi.org/10.1177/02783649241273668},
	doi = {10.1177/02783649241273668},
	language = {EN},
	urldate = {2025-08-28},
	journal = {The International Journal of Robotics Research},
	author = {Chi, Cheng and Xu, Zhenjia and Feng, Siyuan and Cousineau, Eric and Du, Yilun and Burchfiel, Benjamin and Tedrake, Russ and Song, Shuran},
	month = oct,
	year = {2024},
	note = {Publisher: SAGE Publications Ltd STM},
	pages = {02783649241273668},
}

@inproceedings{janner_planning_2022,
	title = {Planning with {Diffusion} for {Flexible} {Behavior} {Synthesis}},
	url = {https://api.semanticscholar.org/CorpusID:248965046},
	booktitle = {International {Conference} on {Machine} {Learning}},
	author = {Janner, Michael and Du, Yilun and Tenenbaum, Joshua B. and Levine, Sergey},
	year = {2022},
}

@inproceedings{dantec_whole_2021,
	title = {Whole {Body} {Model} {Predictive} {Control} with a {Memory} of {Motion}: {Experiments} on a {Torque}-{Controlled} {Talos}},
	doi = {10.1109/ICRA48506.2021.9562092},
	booktitle = {2021 {IEEE} {International} {Conference} on {Robotics} and {Automation} ({ICRA})},
	publisher = {IEEE},
	author = {Dantec, Ewen and Budhiraja, Rohan and Roig, Adria and Lembono, Teguh and Saurel, Guilhem and Stasse, Olivier and Fernbach, Pierre and Tonneau, Steve and Vijayakumar, Sethu and Calinon, Sylvain and Taix, Michel and Mansard, Nicolas},
	year = {2021},
	pages = {8975--8981},
}

@inproceedings{mansard_using_2018,
	title = {Using a {Memory} of {Motion} to {Efficiently} {Warm}-{Start} a {Nonlinear} {Predictive} {Controller}},
	url = {https://ieeexplore.ieee.org/document/8463154},
	doi = {10.1109/ICRA.2018.8463154},
	urldate = {2025-08-29},
	booktitle = {2018 {IEEE} {International} {Conference} on {Robotics} and {Automation} ({ICRA})},
	author = {Mansard, N. and DelPrete, A. and Geisert, M. and Tonneau, S. and Stasse, O.},
	month = may,
	year = {2018},
	note = {ISSN: 2577-087X},
	pages = {2986--2993},
}

@inproceedings{jetchev_trajectory_2009,
	address = {New York, NY, USA},
	series = {{ICML} '09},
	title = {Trajectory prediction: learning to map situations to robot trajectories},
	isbn = {978-1-60558-516-1},
	url = {https://doi.org/10.1145/1553374.1553433},
	doi = {10.1145/1553374.1553433},
	booktitle = {Proceedings of the 26th {Annual} {International} {Conference} on {Machine} {Learning}},
	publisher = {Association for Computing Machinery},
	author = {Jetchev, Nikolay and Toussaint, Marc},
	year = {2009},
	note = {event-place: Montreal, Quebec, Canada},
	pages = {449--456},
}

@inproceedings{lembono_learning_2020,
	title = {Learning {How} to {Walk}: {Warm}-starting {Optimal} {Control} {Solver} with {Memory} of {Motion}},
	shorttitle = {Learning {How} to {Walk}},
	url = {https://ieeexplore.ieee.org/document/9196727},
	doi = {10.1109/ICRA40945.2020.9196727},
	urldate = {2025-08-29},
	booktitle = {2020 {IEEE} {International} {Conference} on {Robotics} and {Automation} ({ICRA})},
	author = {Lembono, Teguh Santoso and Mastalli, Carlos and Fernbach, Pierre and Mansard, Nicolas and Calinon, Sylvain},
	month = may,
	year = {2020},
	note = {ISSN: 2577-087X},
	pages = {1357--1363},
}

@article{makulavicius_industrial_2023,
	title = {Industrial {Robots} in {Mechanical} {Machining}: {Perspectives} and {Limitations}},
	volume = {12},
	issn = {2218-6581},
	url = {https://www.mdpi.com/2218-6581/12/6/160},
	doi = {10.3390/robotics12060160},
	number = {6},
	journal = {Robotics},
	author = {Makulavičius, Mantas and Petkevičius, Sigitas and Rožėnė, Justė and Dzedzickis, Andrius and Bučinskas, Vytautas},
	year = {2023},
}

@misc{peebles_scalable_2023,
	title = {Scalable {Diffusion} {Models} with {Transformers}},
	url = {http://arxiv.org/abs/2212.09748},
	doi = {10.48550/arXiv.2212.09748},
	language = {en},
	urldate = {2025-07-05},
	publisher = {arXiv},
	author = {Peebles, William and Xie, Saining},
	month = mar,
	year = {2023},
	note = {arXiv:2212.09748 [cs]},
}

@misc{haffemayer_collision-free_2025,
  title        = {Collision-Free Model Predictive Control with Diffusion Model Warm-Starting},
  year         = {2025},
  note         = {Preprint, submitted to RAL},
  author       = {Haffemayer, Arthur and Chapin, Alexandre and Wojciechowski, Krzysztof and Jordana, Armand and Lamiraux, Florent and Petr{\'i}k, Vladim{\'i}r and Mansard, Nicolas},
}

@inproceedings{ho_denoising_2020,
	address = {Red Hook, NY, USA},
	series = {{NIPS} '20},
	title = {Denoising diffusion probabilistic models},
	isbn = {978-1-7138-2954-6},
	booktitle = {Proceedings of the 34th {International} {Conference} on {Neural} {Information} {Processing} {Systems}},
	publisher = {Curran Associates Inc.},
	author = {Ho, Jonathan and Jain, Ajay and Abbeel, Pieter},
	year = {2020},
	note = {event-place: Vancouver, BC, Canada},
}

@inproceedings{mirabel2016hpp,
title={HPP: A new software for constrained motion planning},
author={Mirabel, Joseph and Tonneau, Steve and Fernbach, Pierre and Sepp{"a}l{"a}, Anna-Kaarina and Campana, Mylene and Mansard, Nicolas and Lamiraux, Florent},
booktitle={2016 IEEE/RSJ International Conference on Intelligent Robots and Systems (IROS)},
pages={383--389},
year={2016},
organization={IEEE}
}

@article{huang2025prrtc,
  title={prrtc: Gpu-parallel rrt-connect for fast, consistent, and low-cost motion planning},
  author={Huang, Chih H and Jadhav, Pranav and Plancher, Brian and Kingston, Zachary},
  journal={arXiv preprint arXiv:2503.06757},
  year={2025}
}

@inproceedings{bialkowski2011massively,
  title={Massively parallelizing the RRT and the RRT},
  author={Bialkowski, Joshua and Karaman, Sertac and Frazzoli, Emilio},
  booktitle={2011 IEEE/RSJ International Conference on Intelligent Robots and Systems},
  pages={3513--3518},
  year={2011},
  organization={IEEE}
}

@inproceedings{maric2019fast,
title={Fast manipulability maximization using continuous-time trajectory optimization},
author={Mari{'c}, Filip and Limoyo, Oliver and Petrovi{'c}, Luka and Ablett, Trevor and Petrovi{'c}, Ivan and Kelly, Jonathan},
booktitle={2019 IEEE/RSJ International Conference on Intelligent Robots and Systems (IROS)},
pages={8258--8264},
year={2019},
organization={IEEE}
}

@inproceedings{kennel2021manipulability,
  title={Manipulability optimization for multi-arm teleoperation},
  author={Kennel-Maushart, Florian and Poranne, Roi and Coros, Stelian},
  booktitle={2021 IEEE International Conference on Robotics and Automation (ICRA)},
  pages={3956--3962},
  year={2021},
  organization={IEEE}
}

@article{yoshikawa1985manipulability,
  title={Manipulability of robotic mechanisms},
  author={Yoshikawa, Tsuneo},
  journal={The international journal of Robotics Research},
  volume={4},
  number={2},
  pages={3--9},
  year={1985},
  publisher={Sage Publications Sage CA: Thousand Oaks, CA}
}

@article{frison2020hpipm,
  title        = {HPIPM: a high-performance quadratic programming framework for model predictive control},
  author       = {Frison, Gianluca and Diehl, Moritz},
  journal      = {arXiv preprint arXiv:2003.02547},
  year         = {2020},
  doi          = {10.48550/arXiv.2003.02547},
  url          = {https://arxiv.org/abs/2003.02547}
}

\end{document}